\documentclass[conference]{IEEEtran}  

\usepackage{amsmath}
\usepackage{amssymb}
\usepackage{amsfonts}
\usepackage{latexsym}
\usepackage{lipsum}
\usepackage{multicol}
\usepackage{accents}
\usepackage{harpoon}
\usepackage{wrapfig}
\usepackage{tabularx}
\usepackage{euscript}
\usepackage{color}
\usepackage{xcolor}
\usepackage{graphicx}
\usepackage{float}
\usepackage[caption = false]{subfig}	
\usepackage{epsfig}
\usepackage{hyperref}
\usepackage{setspace}
\usepackage{fancyhdr}
\usepackage{textcomp}
\usepackage{cite}
\usepackage{epstopdf}
\usepackage{subfig}
\usepackage{verbatim}
\usepackage[left=0.75in,top=1.444in,right=0.458in,bottom=0.75in]{geometry} 
\DeclareGraphicsExtensions{.pdf,.eps,.jpg, .png}
\graphicspath{{Figures/}}

\usepackage{pifont}
\usepackage{latexsym}
\usepackage{psfrag}
\usepackage{stmaryrd}

\usepackage[linesnumbered,ruled,vlined]{algorithm2e}

\title{
A Multi-Robotic System for Environmental Cleaning
}
\vspace{-15pt}

\author{\IEEEauthorblockN{Chuong Le}
\IEEEauthorblockA{School of Electrical Engineering\\ and Computer Science\\
University of Oklahoma, Norman\\
Email: khuechuong@gmail.com}
\and
\IEEEauthorblockN{Huy Xuan Pham}
\IEEEauthorblockA{ School of Computer Science and\\ Engineering\\ University of Nevada, Reno \\ 
Email: huy.pham@nevada.unr.edu}
\and
\IEEEauthorblockN{Hung Manh La}
\IEEEauthorblockA{Assistant Professor, ARA Lab Director\\
School of Electrical and Computer Engineering\\
University of Nevada, Reno\\
Email: hla@unr.edu}}

\begin{document}

\maketitle
\thispagestyle{empty}
\pagestyle{empty}

\begin{abstract}
There is a lot of waste in an industrial environment that could cause harmful effects to both the products and the workers resulting in product defects, itchy eyes or chronic obstructive pulmonary disease, etc. While automative cleaning robots could be used, the environment is often too big for one robot to clean alone in addition to the fact that it does not have adequate stored dirt capacity. We present a multi-robotic dirt cleaning system algorithm for multiple automatic iRobot Creates teaming to efficiently clean an environment. Moreover, since some spaces in the environment are clean while others are dirty, our multi-robotic system possesses a path planning algorithm to allow  the robot team to clean efficiently by spending more time on the area with higher dirt level. Overall, our multi-robotic system outperforms the single robot system in time efficiency while having almost the same total battery usage and cleaning efficiency result.

\end{abstract}


\section{Introduction}\label{S.1}

In an industrial environment, such as a factory or a warehouse, the work conditions are harsher than normal with wastes - hazardous or non-hazardous lying around, which the workers are exposed to every day. Since hazardous wastes cause visible damages, they are often more carefully handled than non-hazardous ones like metal dusts or ashes, the damages of which are long-term and not always visible. One of the damages that non-hazardous wastes can cause is product defects, sabotaging the purpose of the product and making the producer look bad. In addition, workers working in such an environment may have chronic medical issues such as skin problems, sinusitis, and eye problems. In particular, if inhaled, the dust may cause such symptoms as coughing  or breathing issues and, in more serious cases, could damage the worker's lungs or other organs \cite{doi:10.1093/bmb/ldg034}. If they try cleaning the environment by sweeping the dust, it will go into the air and create a higher chance of the workers inhaling it.

	In this situation, using a dirt cleaning robot, such as the iRobot Roomba, would be a logical solution. Commercial automatic cleaning robots have become very popular nowadays because of their ability to clean autonomously. However, one of the earliest problems with it is that their cleaning process is completely random, and there is no guarantee that they would be able to clean all the space. Recently, though, there are new models of iRobot that could map an entire environment and systematically clean it. 
	
	Even though mapping and cleaning every inch of the environment indiscriminately can be an effective method, it lacks the ability to plan the best path as some areas of the map might be dirtier while others remain clean. Having a dirt model to estimate the amount of dirt in each small area of the environment will help create a more efficient path of cleaning for the robot to follow and, thereby, will increase cleaning efficiency.
	
\begin{figure}[htb!]
\centering
\includegraphics[width=1\columnwidth]{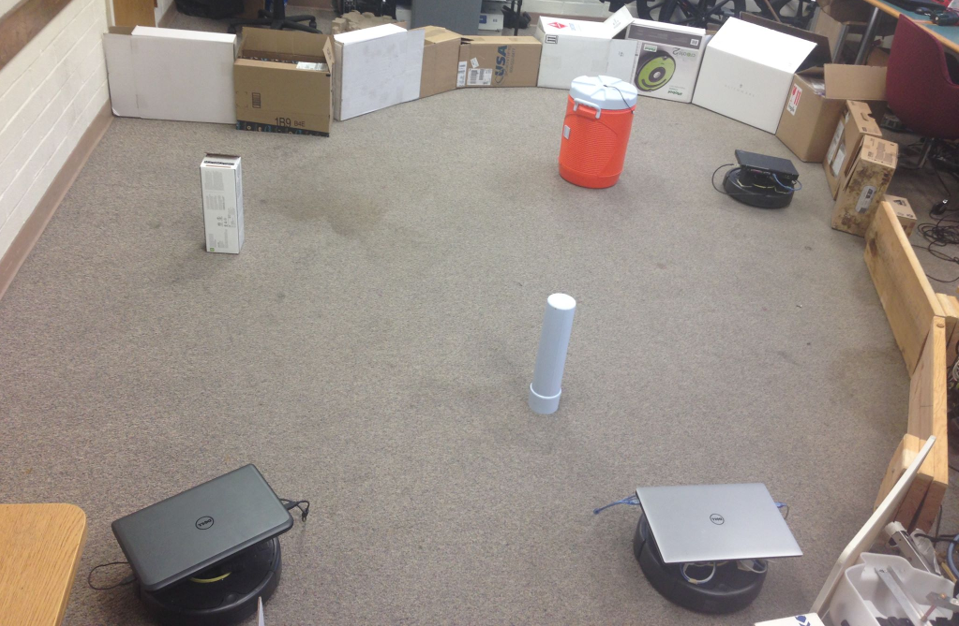}
  \caption{The three iRobot Create 2 that we used in the experiment.  We mounted a hokuyo laser scanner on top of each robot for mapping and localization, and a laptop computer for algorithm processing. In addition, the equipped dirt detection sensor of the iRobot Create will be used to collect environment's dirt readings at all location.}
  \label{irobot image}
\vspace{-0pt}
\end{figure}

Although most automatic cleaning robots could systematically clean a whole environment, an industrial environment is much too large for one robot to clean by itself in addition to the fact that the amount of dirt will exceed the dirt stored capacity of one robot. In this paper, we will be using multiple iRobot Create 2 to build a robotic cleaning team because of its low cost and its programmable platform ~\cite{Tribelhorn2007EvaluatingTR}. In addition, we will use the hokuyo laser scanner, not shown in Figure 1, and the dirt sensor that is already equipped in the iRobot Create 2 to create a dirt map of the environment. Then we will apply the multi-robotic system, where all robots share the same map, know their positions in it, and effectively communicate with each other, to clean an entire industrial environment efficiently.

This paper is presented in the following order: After the related work, we will be discussing the method of mapping and localizing method on Simultaneous Localization and Mapping (SLAM). After that, we will do a brief overview of the cell-wise Poisson process mapping. Then, we will present the traveling salesman problem (TSP) based  path planning algorithm  for the multi-robot system to create an efficient path planning for each robot in the team. Finally, we will present the result of our experiment, evaluate our work and discuss the future usage of our results in the conclusion. 

\section{Related Work }\label{S.2}
Multi-robotic system has many benefits, mainly because it can accomplish task that single robot could not. Prime examples of its benefits include ~\cite{Corts2002CoverageCF, LA2012996, doi:10.1177/1729881418773874, 7852281, Nikitenko2014MultirobotSF, 6783781, 8331947, 7743469, Nguyen2015CompressiveAC, doi:10.1002/rnc.3687, Nguyen_TCNS2018}. ~\cite{Nikitenko2014MultirobotSF} uses rapidly exploring random tree (RRT) for vacuum robot navigation and path planning for large indoor vacuum cleaning. 
~\cite{doi:10.1177/1729881418773874} by Connell et al. uses the RRT$^{x}$ for path replanning in a non-static environment. The multi-robot algorithm here yields equivalent or better paths and planning time efficient. ~\cite{8331947}
by Pham et al creates a wildfire distribution model using multiple unmanned aerial vehicles (UAVs) to track and predict wildfire spreading. In this case if they were using a single UAVs it would not be able to cover the entire fire and create the wildfire model. Another case of using multiple robots for scalar field mapping \cite{La_SMCA2015, La_Cyber2013} that clearly indicated the benefit of multi-robot teaming rather than a single one.

If the robot is cleaning the same environment again, it would be a waste of time to re-scan the environment. In a paper by Zaman et al. ~\cite {Zaman2011ROSbasedML}, it was shown that laser and odometry data from scanning of  a whole environment using SLAM can be saved into Yet Another Markup Language(YAML) and portable gray-map format (PGM) files using map saver and map server. YAML files shows the image while PGM shows the descriptions of the map. When implementing the Adaptive Monte Carlos Localization (AMCL) with the YAML and PGM file, the robot can localize itself inside the saved map ~\cite{Thrun:2005:PR:1121596}.

One of the work used to navigate efficient path planning and navigation is the cell-wise Poisson process that used equations from the homogeneous Poisson process and the maximum likelihood estimator to  predict where the dirt will be after allowing the robot to clean through the environment several times. Then it would create a dirt map that predicts the intensity of dirt in each cell and, using the TSP, determine an efficient pathing for the iRobot to clean. The project's benefits include less noise and less energy consumption in the robot, because of the time and work efficiency ~\cite{Hess2013PoissondrivenDM}. The idea in this paper is similar to the Poison cell-wise method. In this paper, we will create a dirt map and apply the multi-robotic system and propose path planning algorithm for multiple robots to clean the environment in an efficient manner.  

\section{Method}\label{S.3}
As mentioned before, we used 3 iRobot Create 2 with a hokuyo laser mounted on top of each and a Dell XPS 15 laptop. We used the iRobot Create due to the fact that it is an affordable platform and has a built-in dirt detection sensor at the bottom inside the suction unit. The iRobot Create uses a piezoelectric sensor, which generates electrical pulses when dirt hits it and gives a measurement of the dirt reading ~\cite{rrrr}. In the experiment, we used cardboard boxes to create an environment for the iRobot Creates to clean and substituted play sand as dirt. The play sand were scattered randomly in the amount of 10 grams per cycle.

\subsection{Mapping and Localization}
To build a map, we used the mounted hokuyo laser to scan the environment and implemented SLAM Gmapping ~\cite{WinNT} to convert the laser data into a 2D map. Once the environment map is completely built, we saved it and its data to send to the other two iRobots. After each robot received the map, they will localize themselves by using the Adaptive Monte Carlos Localization (AMCL) ~\cite{Zaman2011ROSbasedML, nav},  so they will know their positioning in the environment.

\subsection{Dirt Map}
\begin{figure}
\centering
\includegraphics[width=1\columnwidth]{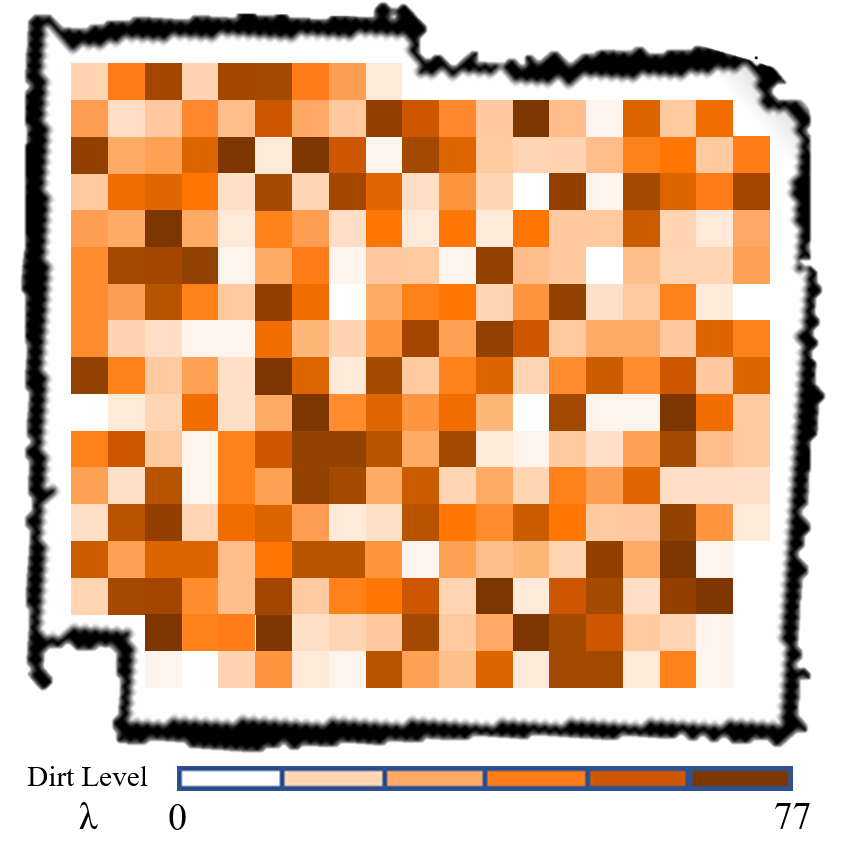}
  \caption{The dirt map visualization of our environment after collecting the data. The colors of the square represents the dirtiness of the area. For example, white squares represents the clean areas and as the squares get darker, the dirtier the square area is.}
  \label{dirt }
\vspace{-10pt}
\end{figure}

For mapping the dirt level, we divided the environment into small square cells and have the iRobots clean the environment several times so that the dirt sensor can collect enough dirt data. Then to predict the dirt level of each cell, we use the homogeneous Poisson process used by Hess et al ~\cite{Hess2013PoissondrivenDM}. The expected dirt level or $\lambda$ for each cell $c$ in the $[s, t]$ interval is

\begin{equation}
E[N^{c}(t)] = \frac{(t-s)}{\sum (t_{i} - t_{i-1})}\sum_{i=1}^{n} k^{c}_{i}, 
\end{equation}

where $s$ is the latest cleaning. Every time the iRobot past through each cell $c$, it takes the dirt reading $k^{c}_{i}$, cleans the cell, and saves the time $t_{i}$. Therefore, $\sum_{i=1}^{n} k^{c}_{i} $ stands for the sum of all dirt reading of cell $c$. In the dirt map of Figure 3, the color of the squares represents the dirt level of the area.

\subsection{Multi-robotic Co-operative Cleaning Algorithm}
Before we implement the algorithm, we took the total dirt level $\lambda_{total}$ of the entire dirt map. Then we divided it by three (since there are three cleaning iRobots) to determine $\lambda_{s}$, the total amount of dirt in each divided cleaning space.

We then created a partition algorithm to divide the dirt map into three separated regions and randomly assigned each iRobot to one region. The regions with dirtier cells will be smaller, and the cleaner regions will be larger to make sure the robots have relatively equal dirt cleaning and therefore similar time. That way, no iRobot will stand around while the others are working in addition to the fact that it will prevent the iRobots from colliding into each other since each has its own space to clean.

\begin{algorithm*}
\SetAlgoLined
Initialize $r$, the number of robots available.

Initialize $\lambda_{total}$, the total dirt level of the entire map.

Initialize $\lambda_{actual}$, the actual total dirt level of a single cleaning space and set it to 0.

Initialize $\lambda_{s}$, set it equals to $\lambda_{total} / r$.

Initialize a vertex matrix $M$.

Initialize $V^{start}$, the starting Vertex and set it to null.

Initialize a vertex array $Q$, of the dirt model where each vertex is a cell:

Vertex $V^{i}$ has a coordinate $(x, y)$, a dirt level $\lambda$, a boolean $visit$ that is $true$ when $V^{start}$ is connected or pass through it and an array of vertex it is connected to.

Initialize integer $n$ and set it to null.

\For {each V in Q} {
	Set $V^{i}_{y} - V^{i}_{x}$ to Integer $diff$. 
	
	\If {$(V^{start} == null)$ or $(diff > n)$}
	{
		Set $diff$ to $n$. Set $V^{i}$ to $V^{start}$.
	}
}

Initialize $counter$ to keep track of the number of maps created and set it equals to 0.

Initialize $counter_2$ to keep track of maps those $\lambda_{actual}$ is not equals to $lambda_{s}$.

Inititialize $r$ and set it equals to 0.

Inititialize $c$ and set it equals to 0.

\While {$counter < r$}
{
	\If {($V^{start}$ is connected to a vertical $V^{i}$) and ($V^{start}$ has not pass through $V^{i}$)}
	{
		Add $V^{i}$ to row $r$ and column $c$ of the matrix $M$ and set $V^{i}_{visit}$ = $true$.
		Add $\lambda_{V^{i}}$ to $\lambda_{actual}$. Set $V^{i}$ to  $V^{start}$.
		Increment $c$. 
	}
	
	\ElseIf {($V^{start}$ is connected to a horizontal $V^{i}$) and ($V^{start}$ has not pass through $V^{i}$)}
	{
		Add $V^{i}$ to row $r$ of the matrix $M$ and set $V^{i}_{visit}$ = $true$.
		Add $\lambda_{V^{i}}$ to $\lambda_{actual}$. Set $V^{i}$ to  $V^{start}$.		
		Increment $c$.
	}
	
	\Else
	{
		Decrement $c$.
		Set $M_{(r,c)}$ to $V^{start}$. Try a different path with $V^{start}$ connecting to $V^{i}_{visit}$ = $false$.
		
		\While {path is not found}
		{
			Decrement $c$. Try a different path with $V^{start}$ with $V^{i}_visit$ = $false$.
			
			\If {path is found}
			{
				Increment $c$ back up.		
			}
		}	
	}

\If {$\lambda_{actual} > \lambda_{s}$}
	{
		\If {$counter_2 / 2$ not equals 0}
		{
			Increment $r$.
			Increment $counter$.	
			Increment $counter_2$.
			Set $\lambda_{actual}$ back to 0.
		}
		\Else
		{
			Take the last $V^{i}$ added to row $r$ and set $V^{i}_{visit}$ to $false$.		
		
			Increment $r$.
			Increment $counter$.	
			Increment $counter_2$.
			Set $\lambda_{actual}$ back to 0.
		}
	}
\ElseIf {$\lambda_{actual} > \lambda_{s}$}
	{
		Increment $r$.
		Increment $counter$.	
		Set $\lambda_{actual}$ back to 0.
	}
}

$return$ matrix $M$
\caption{Multi-Robotic Dirt Cleaning Algorithm}
\label{main algorithm 2}
\end{algorithm*}

\begin{figure}[htb!]
\centering
\includegraphics[width=1\columnwidth]{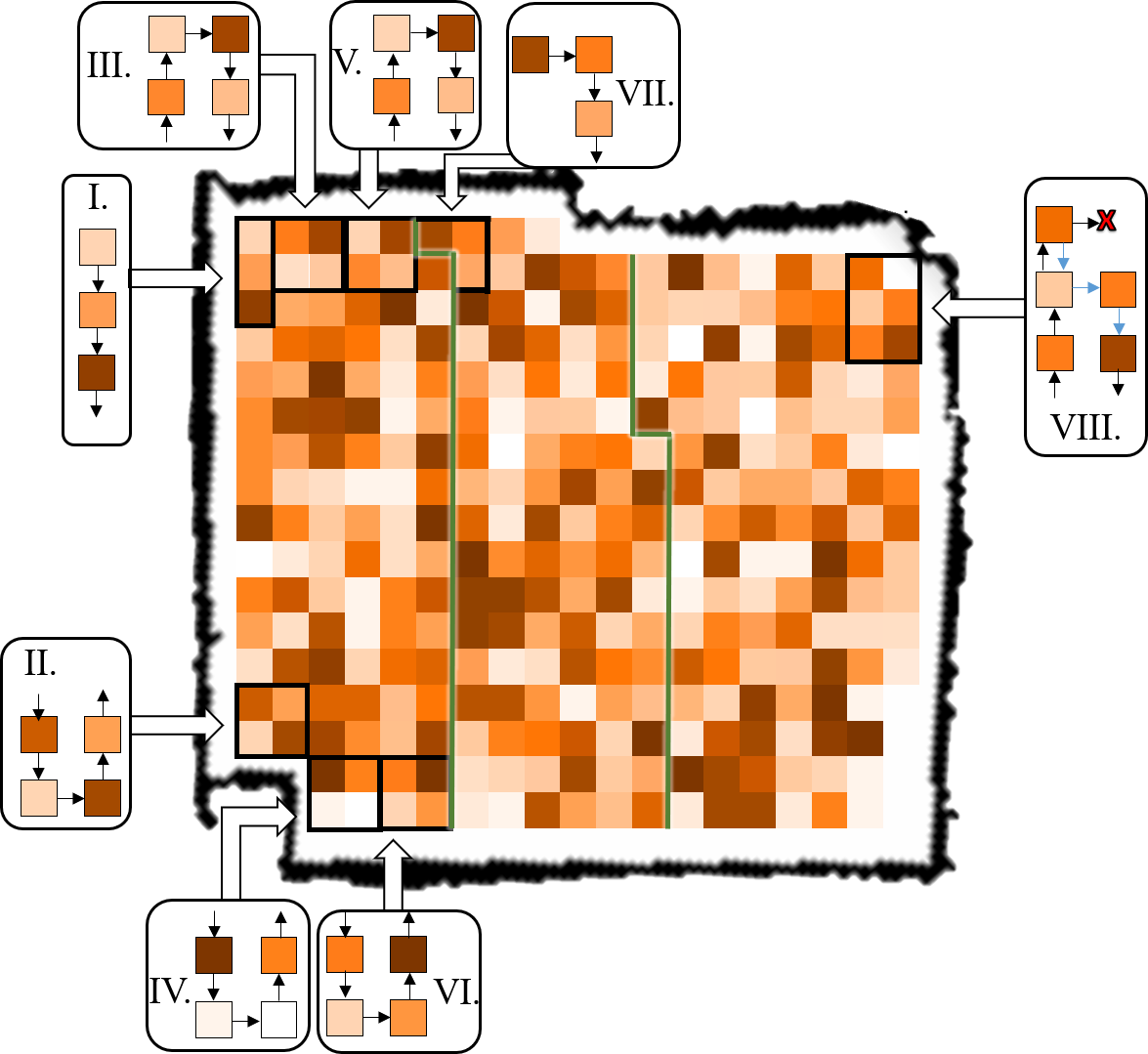}
  \caption{In this dirt model, $\lambda_{total}$ is 11382 and $\lambda_{s}$ is 3794. In step I, we started in the upper left corner (rule 1) and went down (rule 2). In step II,  there is no more vertical vertex left in our path  we turned right  and immediately go upwards (rule 2). Step III, there is no more vertical vertex left in our path so we turned right and go downwards (rule 2). We repeat the process in step IV, V, and VI. The green line represents where we stop and disconnect the vertex we have pass through from the entire map (rule 4) when $\lambda^{0}_{actual}$ = 3808. The reason is our adding rule states that when $\lambda^{i}_{actual}$ adding vertex does not equals to $\lambda_{s}$, we take the $\lambda^{i}_{actual} > \lambda_{s}$ but only if before adding the last vertex, $\lambda^{i}_{actual}$ < $\lambda_{s}$. This is the case since before it reached 3808, it was 3745. The process after step VII is the same as step III, IV, and V repeating until $\lambda^{1}_{actual}$ reached 3748. The reason why it is less than $\lambda_{s}$ is because since the first one is bigger than $\lambda_{s}$ the second one should be less than to maintain the balance. The next green line represent the second cleaning space disconnect from the map. Since that green line is the last process, we do not need to divide anymore (rule 4). Even though step VIII is unnecessary since we already have cleaning space, it shows rule (5).}
  \label{mul}
\vspace{-10pt}
\end{figure}

In this algorithm, we utilize graph data structure by turning each square cell in the dirt map into a vertex. We then start finding the first cleaning space by connecting squares next to, below, and above each other with edges while adding $\lambda$ of vertex we passed to $\lambda^{i}_{actual}$, the actually total dirt level of the cleaning space $i$, until the total dirt level of the square is equal to $\lambda_{s}$. However, real life scenarios of this case often adds up to be either less than or bigger than $\lambda_{s}$. So we will accept that if $\lambda^{i}_{actual}$ will not add up to $\lambda_{s}$, then we will accept $\lambda^{i}_{actual}$ that before adding the next vertex is smaller than $\lambda_{s}$ but bigger than $\lambda_{s}$ after adding the vertex. For the next $\lambda^{i}_{actual}$ that will not add up to $\lambda_{s}$, we will accept the $\lambda$ that after adding the next vertex is bigger than $\lambda_{s}$. We repeat that process until we reach the last cleaning space. So when not equals to $\lambda_{s}$, $\lambda^{i}_{actual}$ will follow a loop of bigger than $\lambda_{s}$, then smaller than $\lambda_{s}$. That way, the total $\lambda$ of each cleaning space will be balance because if $\lambda^{i}_{actual}$ of each space is bigger than $\lambda_{s}$, the $\lambda^{3}_{actual}$, or the $\lambda$ of the last cleaning space, will be a lot smaller than the average cleaning space. 

Here are the rules of connecting squares as shown in Algorithm 1: (1) The ideal starting vertex is the one with the least edges (usually one) and if there are multiples, is usually in the top left corner. In addition, it should try to start at the vertex where it left off. If that turns out to not be good, try another starting point. (2) The edge connection process prioritize vertical pathing and moving towards vertex that hasn't been touched. The one exception when we don't go vertical is when there is no vertical edges left to connect. (3) We usually do not return to the vertex we already visited, the one exception would be if there are on other path available and the vertex's $\lambda$ will not be added to the total $\lambda$ of the cleaning space since it was already added. (4) After one cleaning space is determined, all of its edges will be cut off from the rest of the map so we can start dividing the map again unless it is the last cutting process. (5) Anytime connecting square hit a snag, return to the last vertex and try a different combination, if that does not work, go back further. This method ensures it will try every combinations of connecting squares.

The process of this algorithm implemented on our dirt model is shown in Figure 3.

\subsection{Path Planning}
\begin{figure}[h]
\centering
\includegraphics[width=1\columnwidth]{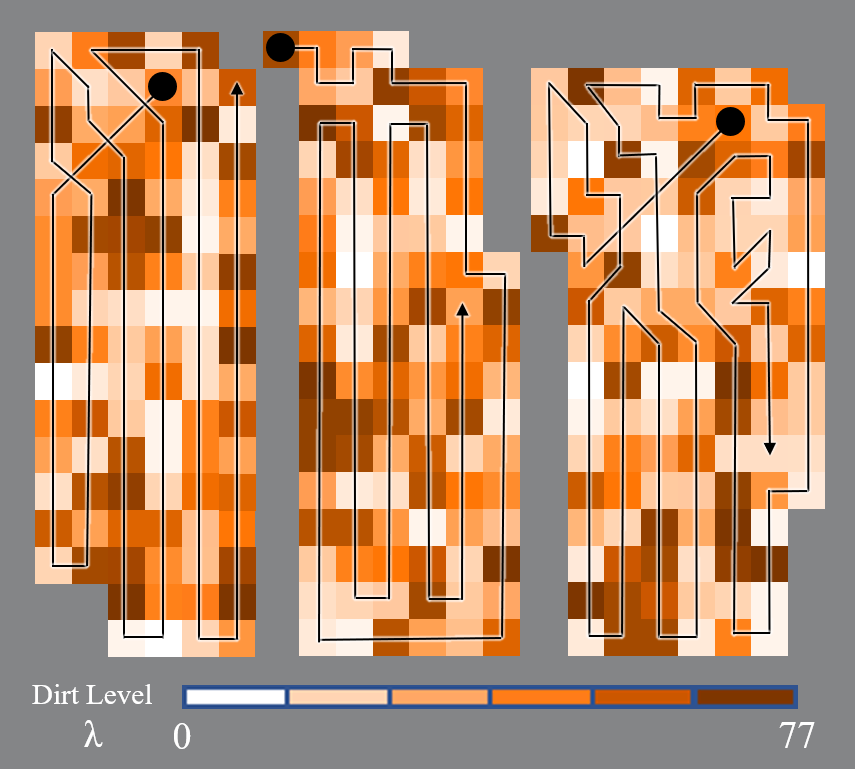}
  \caption{After we assigned cleaning spaces to each iRobots, we treated each space as a traveling salesman problem and solved it to get the pathing. The black circle represents the the starting point while the end of the arrows represents the finish.}
  \label{TSP}
\vspace{-10pt}
\end{figure}

\begin{figure}[h]
\centering
\includegraphics[width=1\columnwidth]{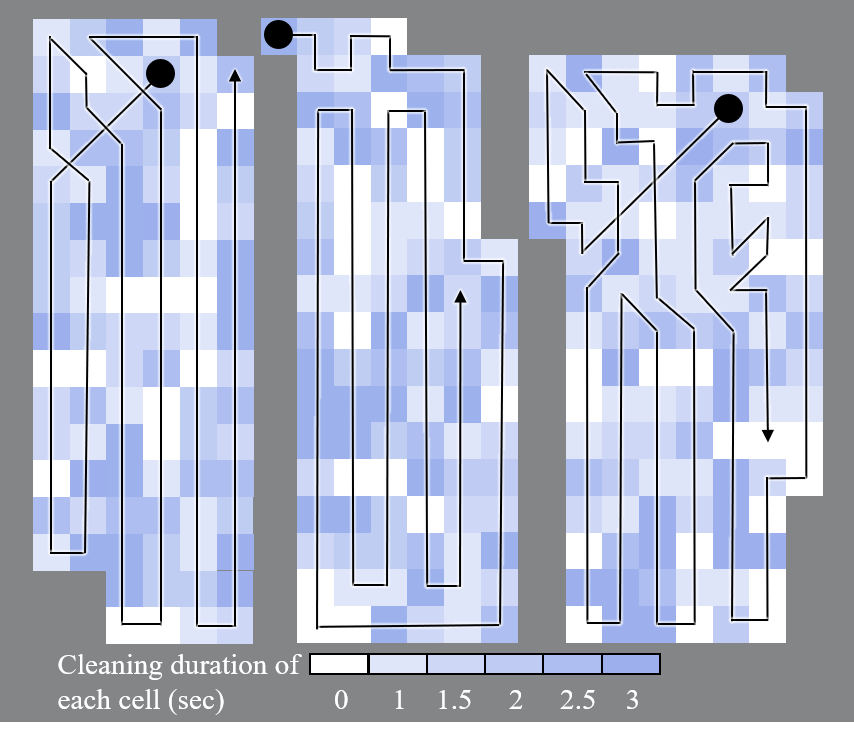}
  \caption{ This is the time map that represents how long the iRobot Create spent cleaning each square cell of the dirt model. The relationship between $\lambda$, dirt level, and duration is as followed: 0-12 = 0 second, 13-26 = 1 second, 27-39 = 1.5 seconds, 40-51 = 2 seconds, 52- 64 = 2.5 seconds, 65-77 = 3 seconds.}
  \label{time}
\vspace{-10pt}
\end{figure}

\begin{figure}[h]
\centering
\includegraphics[width=1\columnwidth]{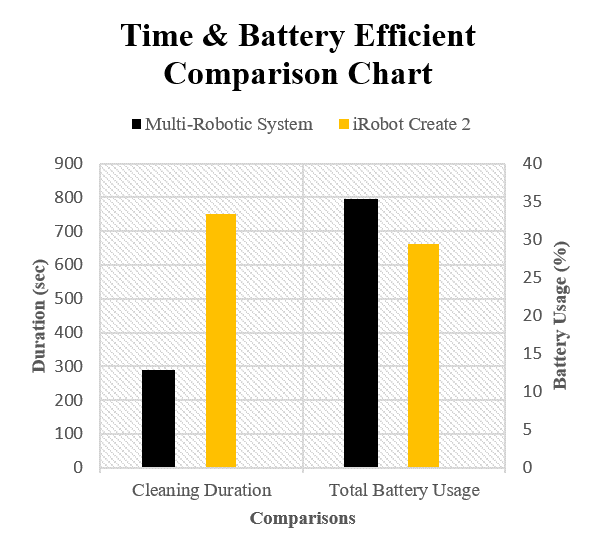}
  \caption{ In the efficiency graph, the left represents the cleaning duration and the right represents total battery usage comparison between three iRobot Create using our multi-robotic (black) and one iRobot Create 2 built-in cleaning system (yellow).}
  \label{result}
\vspace{-10pt}
\end{figure}

Each iRobot will have its own pathing in each of its respective cleaning space. We will treat path planning as a TSP. We utilize TSP because it only goes through each space once and therefore maximizing time efficiency. In each space, first, we create a set of cities $C$, where each city is a square area of the cleaning space and create a set of empty city $P$. Second, we choose an arbitrary starting city from $C$ and add it to $P$. Third, we select a city in $V$ that is the closest to the latest city added to $P$ and add that to $P$. If there are more than one city, then create another set of city identical to $P$ and choose the other city instead. We create more sets to have all pathing possibilities. Fourth, repeat the second step until there are no more cities in $C$. Fifth, we determine which takes the least distance to travel.

Once that is complete, we will repeat the whole process for all starting cities and at the end, compare all of the best path from each starting point and come up with the best pathing. The best pathing for each cleaning space in our dirt model is seen in Figure \ref{TSP}.

Finally, after we have the best pathing, we also determine the amount of time the iRobot will spend cleaning each square. From observing the iRobot's cleaning pattern, we set the cleaning duration based on dirt level as seen in Figure \ref{time}.

\section{Experiment}
\subsection{Testing multi-robotic system versus the built-in system}
In our experiment, we are testing our multi-robotic system against the iRobot Create's cleaning system and evaluate the performance of cleaning time, total battery usage. First, we timed the iRobot Create's cleaning process until it is finished and returned it its home-base. Then we check the battery usage percentage and recorded it. Then, we implemented our multi-robotic system and timed its cleaning process. When all three iRobots stop, we also measure each of its battery usage percentage, added all three together and recorded it. We repeat the process ten times and took the average time and total battery usage. Overall, there were no outliers in either categories - duration and total battery usage - and the spread was relatively small. 
\subsection{Result}

Figure \ref{result} presents the overall result of our experiment. While the total battery usage of all three iRobots is a little higher than the built-in cleaning system of a single iRobot, the total cleaning duration of our multi-robotic system is almost 2/3 less than built-in system of the iRobot Create 2. This experiment shows the time cleaning efficiency of our multi-robotic system while having almost the same battery usage and yielding little to no dirt remaining in the environment.

\section{Conclusion}

In this paper, we present a multi-robotic algorithm along with a dirt model for industrial environment cleaning. After collecting dirt reading data, the dirt model predicts the dirt level for each small area on the environment map. Our algorithm divides the model evenly in terms of total dirt level between three automative iRobot Create 2 and creates the best pathing for each iRobot as well as cleaning duration of each small area depending on its dirt level. In the future, we plan to make our system viable for non-static environment such as when the map changes as well as collecting more data to update the new areas in the dirt model. In addition, we aim to have each of our iRobot be able to do path replanning in case of a moving obstacle like ~\cite{Connell_SMC2017,doi:10.1177/1729881418773874} and avoiding trapped by convex shaped obstacles by using a combination of rotational force field \cite{Dang_MFI2016} and repulsive artificial potential force field \cite{Woods_SMCA2018}.

\section*{Acknowledgment}
This material is based upon work supported by the National Science Foundation (NSF) \#IIS-1757929. The views, opinions, findings and conclusions reflected in this publication are solely those of the authors and do not represent the official policy or position of the NSF.

\addtolength{\textheight}{-12cm}   








\bibliographystyle{IEEEtran}
\bibliography{bibliography}

\end{document}